\documentclass[sigconf]{acmart}
\AtBeginDocument{%
 }

\usepackage{algorithm}
\usepackage{algpseudocode}

\setcopyright{acmlicensed}
\copyrightyear{2027}
\acmYear{2027}
\acmDOI{XXXXXXX.XXXXXXX}
\acmConference[KDD '27]{The 33rd ACM SIGKDD Conference on Knowledge Discovery and Data Mining}{August 2027}{San Jose, CA, USA}
\title{Measuring the Prevalence of Policy-Violating Content with ML-Assisted Sampling and LLM Labeling}

\author{Attila Dobi}
\affiliation{%
 \institution{Pinterest}
 \city{San Francisco}
 \state{CA}
 \country{USA}
}
\email{adobi@pinterest.com}

\author{Aravindh Manickavasagam}
\affiliation{%
 \institution{Pinterest}
 \city{Richmond}
 \state{VA}
 \country{USA}
}
\email{amanickavasagam@pinterest.com}

\author{Xiaohan Yang}
\affiliation{%
 \institution{Pinterest}
 \city{San Francisco}
 \state{CA}
 \country{USA}
}
\email{xiaohanyang@pinterest.com}

\author{Benjamin Thompson}
\affiliation{%
 \institution{Pinterest}
 \city{Los Angeles}
 \state{CA}
 \country{USA}
}
\email{benjaminthompson@pinterest.com}

\author{Faisal Farooq}
\affiliation{%
 \institution{Pinterest}
 \city{Murphy}
 \state{TX}
 \country{USA}
}
\email{ffarooq@pinterest.com}

\begin{abstract}
Content safety teams need metrics that reflect what users actually experienced, not only what was reported.
We study \emph{prevalence}: the fraction of user views (impressions) that went to content violating a given policy on a given day.
Accurate prevalence measurement is challenging because violations are often rare and human labeling is costly, making frequent, platform-representative studies slow.

We present a design-based measurement system that (i) draws daily probability samples from the impression stream using \emph{ML-assisted} weights to concentrate label budget on high-exposure and high-risk content while preserving unbiasedness, (ii) labels sampled items with a multimodal LLM governed by policy prompts and gold-set validation, and (iii) produces design-consistent prevalence estimates with confidence intervals and dashboard drill-downs.
A key design goal is \emph{one global sample with many pivots}: a single daily sample supports prevalence by surface, viewer geography, content age, and other segments without re-sampling.
We detail the statistical estimators, variance and confidence interval construction, a sensitivity analysis of how label error propagates into prevalence estimates, practical guidance for configuring sampling hyperparameters, and an engineering workflow that makes the system configurable across policies.
The system has run in daily production for over a year across multiple policy areas; because absolute prevalence levels for sensitive policies are confidential, we report production evidence in relative terms and include a fully specified synthetic simulation appendix illustrating sampling efficiency.
\end{abstract}



\begin{document}
\maketitle

\section{Introduction}
User-generated content platforms need fast, repeatable feedback loops to detect emerging harms and evaluate mitigations.
Operationally, metric refresh must be frequent enough to surface new patterns, support rapid root-cause analysis, and measure whether interventions and product changes move safety outcomes in the intended direction.

However, a reports-only view is incomplete since many harms are under-reported, some users do not report content they seek, and rare categories yield too few reports for statistically powered monitoring.
\emph{Prevalence} complements reporting by estimating the share of user experiences exposed to policy-violating content, enabling guardrails, trend monitoring, goal-setting, and intervention evaluation \cite{pinterest_radar_2025}; such exposure-based measurement is increasingly expected under transparency regimes like the EU Digital Services Act \cite{dsa2022}.
Because prevalence is computed from probability samples, it also supports drill-downs that help teams localize changes (e.g., by surface, geography, or content age) without re-running bespoke studies.

Historically, prevalence studies relied on human review and were conducted infrequently due to significant cost and latency. Recent multimodal LLMs fundamentally shift this operating point: by bulk-labeling samples with an LLM using subject-matter-expert (SME) reviewed prompts and ongoing gold-set validation, we reduce end-to-end labeling latency by $\sim$15$\times$ (relative to a human-only review workflow) and operational cost by $>10\times$ at comparable decision quality. This efficiency makes daily, platform-representative measurement feasible at scale \cite{pinterest_radar_2025}.

This paper describes an approach that enables daily measurement by combining (i) ML-assisted probability sampling from impression logs and (ii) LLM-assisted labeling with ongoing decision-quality monitoring.
While production enforcement models provide auxiliary risk scores, they are not used as labels and do not gate inclusion; design-based reweighting removes sampling ``lensing'' so estimates remain comparable even as enforcement thresholds evolve.

In production at Pinterest, we run this pipeline daily (for over a year) across multiple policy areas, supporting daily monitoring, OKR-style goal tracking, and automated alerting. Each run publishes prevalence estimates with 95\% CIs and drill-down dashboards, and pages an on-call owner when weekly shifts exceed policy-specific minimum detectable effect (MDE) thresholds (\S\ref{sec:case}).
Because absolute prevalence levels for sensitive policies are confidential, we report production evidence primarily in relative terms (e.g., sampling-efficiency lift and detectable relative changes) and provide a fully specified synthetic simulation appendix (\S\ref{app:sim_pseudocode}) to enable methodological verification. Across policies, prevalence correlates weakly with confirmed user reports, indicating it provides complementary visibility beyond reporting.

\begin{figure}
 \centering
 \includegraphics[width=0.76\linewidth]{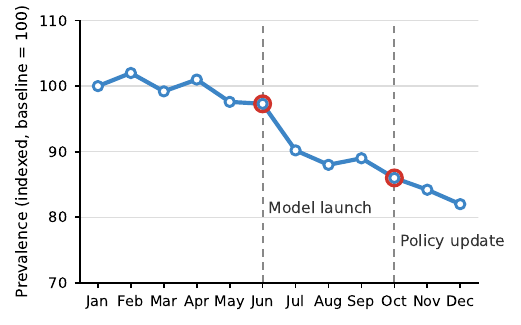}
 \caption{Illustrative prevalence trend (indexed to baseline $=100$) with intervention markers (model launch, policy update). Production dashboards add 95\% CIs, absolute levels, and launch history.}
 \Description{A line chart of indexed prevalence over twelve months with dashed vertical intervention markers labeled model launch and policy update; the line decreases after each intervention.}
 \label{fig:trend}
\end{figure}

\paragraph{Contributions.}
This work makes four contributions:
(1) a design-consistent estimator for daily prevalence of violative views with \emph{single-sample drill-downs} across many segments;
(2) practical uncertainty quantification: CIs, effective sample size, a label-error sensitivity analysis with guidance on activating the Rogan--Gladen correction (including weight-correlated errors), and an anomaly-triage protocol for the upward fluctuations label error induces;
(3) a concrete workflow for configuring the sampling hyperparameters $(\nu,\gamma,\epsilon)$ for a new policy, backed by a simulation-based sensitivity study;
and (4) a configurable engineering workflow that turns a policy definition, an SME-reviewed prompt, and a gold set into a daily measurement pipeline.

\section{Related Work}
Our work draws on classical design-based sampling and estimation.
Probability proportional-to-size (PPS) sampling and the Hansen--Hurwitz estimator provide unbiased (or design-consistent) estimation under with-replacement designs \cite{hansen1943theory,cochran1977sampling}.
For without-replacement designs, Horvitz--Thompson style estimators and their ratio variants are standard tools \cite{horvitz1952generalization}.
Using auxiliary information for efficiency while retaining design-based validity is treated systematically in the model-assisted survey sampling literature \cite{sarndal1992model}; our effective sample size diagnostics follow Kish \cite{kish1965survey}, and weighted reservoir sampling gives an efficient streaming PPSWOR implementation \cite{efraimidis2006weighted}.

Recently, \emph{model-assisted} and \emph{ML-assisted} sampling have been used in industry to improve efficiency when rare events are important but unbiasedness is required.
Meta (Facebook) released an open-source implementation (\texttt{ml\_sampler}) that combines auxiliary model scores with probability sampling for estimation \cite{facebook_mlsampler}.
In the content-safety setting, prevalence-style exposure metrics complement reports by measuring what was actually shown to users and enabling intervention evaluation \cite{pinterest_radar_2025}.

On the labeling side, there is growing evidence that LLMs can serve as high-quality annotators for certain tasks, often matching or exceeding crowd-worker performance when prompts are well specified \cite{gilardi2023chatgpt}.
Industrial deployments of LLM-based policy interpretation are emerging across platforms: Google scaled LLM reviews for ads moderation \cite{qiao2024scaling}, OpenAI described a production pipeline for undesired-content detection \cite{markov2023holistic}, Meta released the Llama Guard safeguards \cite{inan2023llamaguard}, and recent work benchmarks multimodal LLMs against professional human moderators \cite{levi2025aivshuman}.
These systems target review and enforcement; ours is complementary: unbiased, exposure-weighted \emph{measurement} with the LLM as a governed labeling instrument.
Because LLM behavior can drift (model updates, content shift, policy changes), practical deployments require ongoing monitoring against human judgment and gold sets; rubric-based and judge-based evaluation frameworks provide useful patterns \cite{zheng2023judging,hashemi2024llmrubric}, and a companion paper details the decision-quality evaluation framework used to govern the labelers in this system \cite{tian2026dqframework}.

\section{Problem and Metric}
\label{sec:problem}
Fix a reporting day. The population is the set of content units (e.g., Pins) that received at least one impression that day.
For each content unit $j$ we observe its total impressions $C_{j} \ge 0$ from impression logs.
A policy-specific label $Y_{j} \in \{0,1\}$ indicates whether $j$ violates the policy (1) or not (0) under the policy definition in effect on that day.

\paragraph{Notation.}
All quantities are per-day; we suppress the day subscript through \S\ref{sec:uq}, reintroducing it only where time aggregation requires it (\S\ref{sec:case}). Table~\ref{tab:notation} summarizes symbols.

\begin{table}
 \caption{Notation (all quantities per day; day subscript suppressed through \S\ref{sec:uq}).}
 \label{tab:notation}
 \footnotesize
 \begin{tabular}{@{}ll@{}}
 \toprule
 Symbol & Meaning \\
 \midrule
 $j$ & content unit (e.g., a Pin) \\
 $C_{j}$, $C_{jg}$ & impressions of $j$ (total; in segment $g$) \\
 $Y_{j}\in\{0,1\}$ & policy-violation label of $j$ \\
 $s_{j}$ & auxiliary risk score (sampling only, never a label) \\
 $w_{j}$, $p_{j}$ & sampling weight; normalized draw probability \\
 $\nu,\gamma,\epsilon$ & weight hyperparameters, Eq.~\eqref{eq:weight} \\
 $m$; $J_i$ & daily label budget; $i$-th sampled unit \\
 $\theta$, $\theta_{g}$ & exposure-weighted prevalence (overall; segment $g$) \\
 $r$, $f$ & labeler sensitivity (recall); false positive rate \\
 $\mathrm{ESS}$ & Kish effective sample size \\
 \bottomrule
 \end{tabular}
\end{table}

We measure exposure-weighted prevalence:
\begin{equation}
\theta \;=\; \frac{\sum_j C_{j} Y_{j}}{\sum_j C_{j}}.
\label{eq:theta_def}
\end{equation}
This definition emphasizes user experience: a single item can be posted once but receive many impressions, or none.

\subsection{Segments and drill-downs}
For drill-downs, we define segments $g \in \mathcal{G}$ over impressions (e.g., surface=Homefeed/Search, viewer country, content age bucket).
Let $C_{jg}$ be the impressions to content $j$ attributable to segment $g$, with $\sum_g C_{jg} = C_{j}$.
Segment prevalence is
\begin{equation}
\theta_{g} \;=\; \frac{\sum_j C_{jg} Y_{j}}{\sum_j C_{jg}}.
\label{eq:theta_seg}
\end{equation}
A primary design goal is that a \emph{single global sample} supports all $g$.
We achieve this by (i) sampling content units once per day using global weights, and (ii) storing $\{C_{jg}\}_{g \in \mathcal{G}}$ for sampled items so the estimator can be evaluated for any pivot without re-sampling.

\section{ML-Assisted Probability Sampling}
\label{sec:sampling}
Violations may be rare, so naive random sampling can yield too few positive examples for stable daily estimates.
We therefore use auxiliary signals to allocate label budget more efficiently while preserving unbiasedness.

\subsection{Sampling weights}
Each content unit $j$ receives an auxiliary risk score $s_{j} > 0$ from production safety models.
Missing scores are imputed (e.g., to the day's median) to keep fresh content in-frame \cite{pinterest_radar_2025}.
We define a nonnegative weight
\begin{equation}
 w_{j}
 = C_{j}^{\nu}\,\bigl(s_{j}^{\gamma} + \epsilon\bigr),
 \qquad \nu,\gamma \ge 0,\; \epsilon>0.
\label{eq:weight}
\end{equation}
The exponents control efficiency trade-offs (Table~\ref{tab:regimes}): $\gamma=0$ yields impression-weighted sampling and $\gamma=\nu=0$ approximately uniform sampling; $\gamma>0$ shifts mass toward model-high-risk items (raising the sample positive rate) but can reduce effective sample size when score distributions are spiky, while $\nu<1$ softens exposure weighting to control weight variance.

\begin{table}
 \caption{Sampling regimes via parameter choices in \eqref{eq:weight}. The floor $\epsilon>0$ keeps every in-frame unit (including unscored, imputed content) at nonzero probability.}
 \label{tab:regimes}
 \begin{tabular}{ccc}
 \toprule
 $\nu$ & $\gamma$ & Interpretation \\
 \midrule
 0 & 0 & Uniform random over content units \\
 1 & 0 & PPS by impressions (exposure-weighted) \\
 1 & 1 & ML-assisted: impressions $\times$ model score \\
 $\in(0,1)$ & 0 & Softer exposure weighting (variance control) \\
 \bottomrule
 \end{tabular}
\end{table}

We construct a daily sampling distribution over content units:
\begin{equation}
 p_{j} = \frac{w_{j}}{\sum_k w_{k}}.
\label{eq:prob}
\end{equation}
Crucially, $s_{j}$ is used only to prioritize review and does not become a label; the estimator reweights by $1/p_{j}$ so prevalence reflects impressions, not model thresholds.

\subsection{Efficient implementation via weighted reservoir sampling}
At platform scale (billions of impressions/day), the population can contain tens of millions of content units; streaming sampling keeps memory $O(m)$ and avoids materializing the full population.
We implement weighted reservoir sampling using the Efraimidis--Spirakis key trick \cite{efraimidis2006weighted}:
for each unit, draw $U_{j}\sim\mathrm{Uniform}(0,1]$ and compute
\begin{equation}
\kappa_{j} = -\frac{\log(U_{j})}{w_{j}}.
\label{eq:key}
\end{equation}
Keeping the $m$ smallest keys yields a fixed-size PPSWOR sample (equivalently, the $m$ earliest events in an exponential race).
The threshold $\tau$ (the $m$-th smallest key) provides a useful inclusion-probability approximation via $\pi_{j}\approx1-\exp(-w_{j}\tau)$; for small sampling fractions, $\pi_{j}\approx w_{j}/\sum_k w_{k}$.
We persist $\tau$ so these approximations (and conservative variance estimators) are recoverable.
We also support with-replacement multinomial draws from \eqref{eq:prob} (PPSWR), which we use for exposition, noting the PPSWOR alternative in \S\ref{sec:estimation}.

\subsection{Choosing \texorpdfstring{$(\nu,\gamma,\epsilon)$}{(nu, gamma, epsilon)} for a new policy}
\label{sec:tuning}
In response to reviewer feedback, we document the onboarding workflow used to configure the sampling hyperparameters for a new policy metric; a simulation-based sensitivity study appears in Appendix~\ref{app:hyper}.
\begin{enumerate}
\item \textbf{Set operational targets.} Choose the weekly relative MDE the team needs (\S\ref{sec:mde}) and a prior base-rate estimate $\theta_0$ (small pilot or adjacent policy); back out the required daily CI half-width $h^\ast$ via Eq.~\eqref{eq:mde_rel} and the required $\mathrm{ESS}^\ast \approx (1.96)^2\,\theta_0/(h^\ast)^2$ via Eq.~\eqref{eq:sens_ess}.
\item \textbf{Initialize.} Start from $\nu=1$ and $\gamma\in\{0,1\}$; set $\epsilon$ to a small positive value below the observed score floor so unscored or imputed items stay in-frame.
\item \textbf{Pilot both schemes.} Run $\gamma=0$ and $\gamma=1$ for a few days at the intended budget $m$, recording Kish ESS, sample positive count, and realized CI width. The positive count is the binding quantity for rare policies.
\item \textbf{Tune to a constrained objective.} Increase $\gamma$ (or soften $\nu<1$ if exposure weights are heavy-tailed) to maximize expected positive count \emph{subject to} $\mathrm{ESS}\ge\mathrm{ESS}^\ast$; the sweep in Appendix~\ref{app:hyper} shows the trade-off is well behaved near $\gamma\in[0.5,2]$ and degrades sharply beyond, so stop at the smallest $\gamma$ meeting the positive-count target.
\end{enumerate}
Production configurations use $\nu=1$, $\gamma=1$ with policy-specific $m$ (\S\ref{sec:case}); $(\nu,\gamma)$ are revisited when the \S\ref{sec:ess} diagnostics degrade or the auxiliary-score model is refreshed (\S\ref{sec:system}).

\section{Design-Consistent Estimation and Single-Sample Drill-downs}
\label{sec:estimation}
Let the daily sample contain $m$ draws $\{J_i\}_{i=1}^m$.
For draw $i$, we observe $(C_{J_i},\,Y_{J_i})$ and the draw probability $p_{J_i}$.
Define
\begin{equation}
 x_{i} = C_{J_i},\qquad z_{i} = C_{J_i} Y_{J_i}.
\end{equation}

\subsection{Hansen--Hurwitz ratio estimator (PPS with replacement)}
For PPSWR, define the HH mean estimators
\begin{equation}
 \hat{Z} = \frac{1}{m}\sum_{i=1}^m \frac{z_{i}}{p_{J_i}},\qquad
 \hat{X} = \frac{1}{m}\sum_{i=1}^m \frac{x_{i}}{p_{J_i}}.
\end{equation}
The Hansen--Hurwitz (HH) ratio estimator \cite{hansen1943theory} is
\begin{equation}
\hat{\theta}
= \frac{\hat{Z}}{\hat{X}}
= \frac{\sum_{i=1}^m \frac{z_{i}}{p_{J_i}}}{\sum_{i=1}^m \frac{x_{i}}{p_{J_i}}}.
\label{eq:hh_ratio}
\end{equation}
In practice, score drift can show up as wider confidence intervals rather than bias, because the reweighting corrects for the sampling design; \S\ref{sec:eval} and Appendix~\ref{app:hyper} quantify this behavior.

\subsection{Without-replacement alternative}
When using weighted reservoir sampling (PPSWOR), we can use a Horvitz--Thompson/H\'ajek style ratio estimator, approximating the inclusion probabilities $\pi_{j}$ via the stored threshold $\tau$ from \eqref{eq:key} (Poissonized approximation) or using variance estimators that do not require exact $\pi_{j}$.
Empirically, for large populations and moderate $m$, PPSWR and PPSWOR yield similar point estimates, with PPSWOR typically slightly lower variance.

\subsection{Drill-downs from a single sample}
For a segment $g$, define per-draw quantities
\begin{equation}
 x_{ig} = C_{J_ig},\qquad z_{ig} = C_{J_ig}Y_{J_i}.
\end{equation}
Two common cases arise:
\begin{itemize}
\item \textbf{Known segment denominators.} If $D_{g}=\sum_j C_{jg}$ is computed exactly from logs, we estimate only the numerator:
\begin{equation}
\widehat{N}_{g}=\frac{1}{m}\sum_{i=1}^m \frac{z_{ig}}{p_{J_i}},
\qquad
\hat{\theta}_{g} = \widehat{N}_{g}/D_{g}.
\label{eq:seg_known_denom}
\end{equation}
\item \textbf{Sample-based denominators.} If $D_{g}$ is not directly available (e.g., ad-hoc pivots), we use the HH ratio form:
\begin{equation}
\hat{\theta}_{g}
= \frac{\sum_{i=1}^m \frac{z_{ig}}{p_{J_i}}}{\sum_{i=1}^m \frac{x_{ig}}{p_{J_i}}}.
\label{eq:seg_ratio}
\end{equation}
\end{itemize}
The critical engineering step is persisting $\{C_{jg}\}$ for sampled items (and optionally a compact segment key) so dashboards can compute \eqref{eq:seg_known_denom} or \eqref{eq:seg_ratio} on demand.

\section{Uncertainty Quantification}
\label{sec:uq}
We report 95\% confidence intervals (CIs) to convey precision.
For PPSWR, we use a Taylor linearization for the ratio estimator.
Define residuals
\begin{equation}
 e_{i} = \frac{1}{p_{J_i}}\bigl(z_{i} - \hat{\theta} x_{i}\bigr).
\end{equation}
An approximate variance estimator is
\begin{equation}
 \widehat{\mathrm{Var}}(\hat{\theta})
 \approx
 \frac{1}{\hat{X}^2}
 \cdot \frac{1}{m(m-1)}\sum_{i=1}^m \bigl(e_{i}-\bar e\bigr)^2,
 \qquad \bar e = \frac{1}{m}\sum_{i=1}^m e_{i}.
\label{eq:var}
\end{equation}
We report
\begin{equation}
\hat{\theta} \pm 1.96\,\sqrt{\widehat{\mathrm{Var}}(\hat{\theta})}.
\label{eq:ci}
\end{equation}
The same construction applies to each segment by replacing $(x_{i},z_{i})$ with $(x_{ig},z_{ig})$.

\subsection{Effective sample size diagnostics}
\label{sec:ess}
Highly variable weights can reduce precision.
To communicate this effect to metric owners, dashboards report an effective sample size (ESS) proxy \cite{kish1965survey}, e.g.
\begin{equation}
\mathrm{ESS} = \frac{\left(\sum_{i=1}^m a_{i}\right)^2}{\sum_{i=1}^m a_{i}^2},
\qquad a_{i}=\frac{x_{i}}{p_{J_i}},
\end{equation}
along with the sample positive rate and CI width.
This is the standard Kish effective sample size proxy for weighted estimators.
These operational diagnostics feed the tuning workflow of \S\ref{sec:tuning}.

\subsection{Label error: bias, correction, and monitoring}
\label{sec:labelerror}
LLM labels have nonzero error rates: a false positive rate $f$ and a sensitivity (recall) $r$, so the false negative rate is $1-r$.
If error rates are approximately homogeneous across items, the LLM-measured prevalence satisfies
\begin{equation}
\theta^{(L)} \;=\; r\,\theta + f\,(1-\theta),
\qquad
\theta^{(L)}-\theta \;=\; f-(1-r+f)\,\theta.
\label{eq:measured}
\end{equation}
Two operating consequences follow.
First, for rare policies ($\theta\ll 1$) the \emph{relative} bias is approximately $f/\theta-(1-r+f)$: the false positive rate, not recall, dominates.
At $\theta=0.1\%$, even $f=0.1\%$ makes roughly half of all measured positives false detections, a relative error approaching $+100\%$ ($+90\%$ at $r=0.90$), whereas recall loss enters only as a bounded $-(1-r)$ offset (Appendix~\ref{app:label_error}, Figure~\ref{fig:label_bias}); this is why prompt optimization minimizes FPR subject to recall constraints (\S\ref{sec:llm}) and why we monitor $f$ at least as closely as $r$.
Second, if we estimate $(r,f)$ on a contemporaneous validation set, we can correct the prevalence via Rogan--Gladen \cite{rogan1978estimating}:
\begin{equation}
\theta = \frac{\theta^{(L)} - f}{r - f},
\label{eq:rg}
\end{equation}
propagating uncertainty by combining sampling variance \eqref{eq:var} with validation uncertainty on $(r,f)$ via the delta method or bootstrap (Appendix~\ref{app:label_error}).

\paragraph{Weight-correlated label errors.}
If per-item error rates $(r_j,f_j)$ correlate with exposure or risk (e.g., high-risk content is systematically harder), the estimand of $\hat\theta^{(L)}$ involves \emph{impression-weighted} rates $(\bar r,\bar f)$, which convenience validation sets misstate (Appendix~\ref{app:label_error}).
We therefore estimate $(r,f)$ on validation subsamples drawn from the same probability design (mechanism (1) of \S\ref{sec:gating}), making $(\hat r,\hat f)$ design-consistent for the exposure-weighted rates.

\paragraph{When to activate the correction.}
In production we keep the core metric simple: we monitor $(\hat r,\hat f)$ against launch-gated tolerance thresholds (\S\ref{sec:gating}) and report the uncorrected $\theta^{(L)}$, activating \eqref{eq:rg} when $\hat f$ becomes non-negligible relative to the measured level (rule of thumb: $\hat f \gtrsim 0.1\,\hat\theta^{(L)}$) or when monitored LLM-vs-human bias approaches the policy's MDE, so that labeler error cannot masquerade as a real trend.

\paragraph{Upward fluctuations and anomaly triage.}
Equation~\eqref{eq:measured} implies an operational asymmetry: false positives contribute $f(1-\theta)$ while missed recall subtracts only $(1-r)\theta$, so in the rare-event regime label error pushes the measured series \emph{upward}. Anomalous upticks are therefore treated as suspect until triaged.
Beyond the linearized variance \eqref{eq:var}, we bootstrap the day's $m$ draws (resampling and recomputing \eqref{eq:hh_ratio}) for interval estimates robust to heavy-tailed weights, with per-unit influence diagnostics flagging estimates dominated by a few high-weight positives.
When an alert fires (\S\ref{sec:case}), we re-run labeling on the same sampled units and route the highest-influence positives to SME adjudication before the movement is accepted as genuine; if labeler bias is confirmed, we refresh $(\hat r,\hat f)$, apply \eqref{eq:rg}, and backfill the affected days.

\paragraph{What a false positive means here.}
Labeler false positives in this domain are rarely benign content: SME adjudication typically finds borderline material adjacent to the policy line, so an inflated series still tracks degraded user experience.
The metric's contract, however, is strict coherence with the written policy, so such items are treated as measurement error, adjudicated, and used to tighten prompts (\S\ref{sec:case}).

\section{LLM Labeling and Decision Quality Monitoring}
\label{sec:llm}
\begin{figure}
 \centering
 \includegraphics[width=0.62\linewidth]{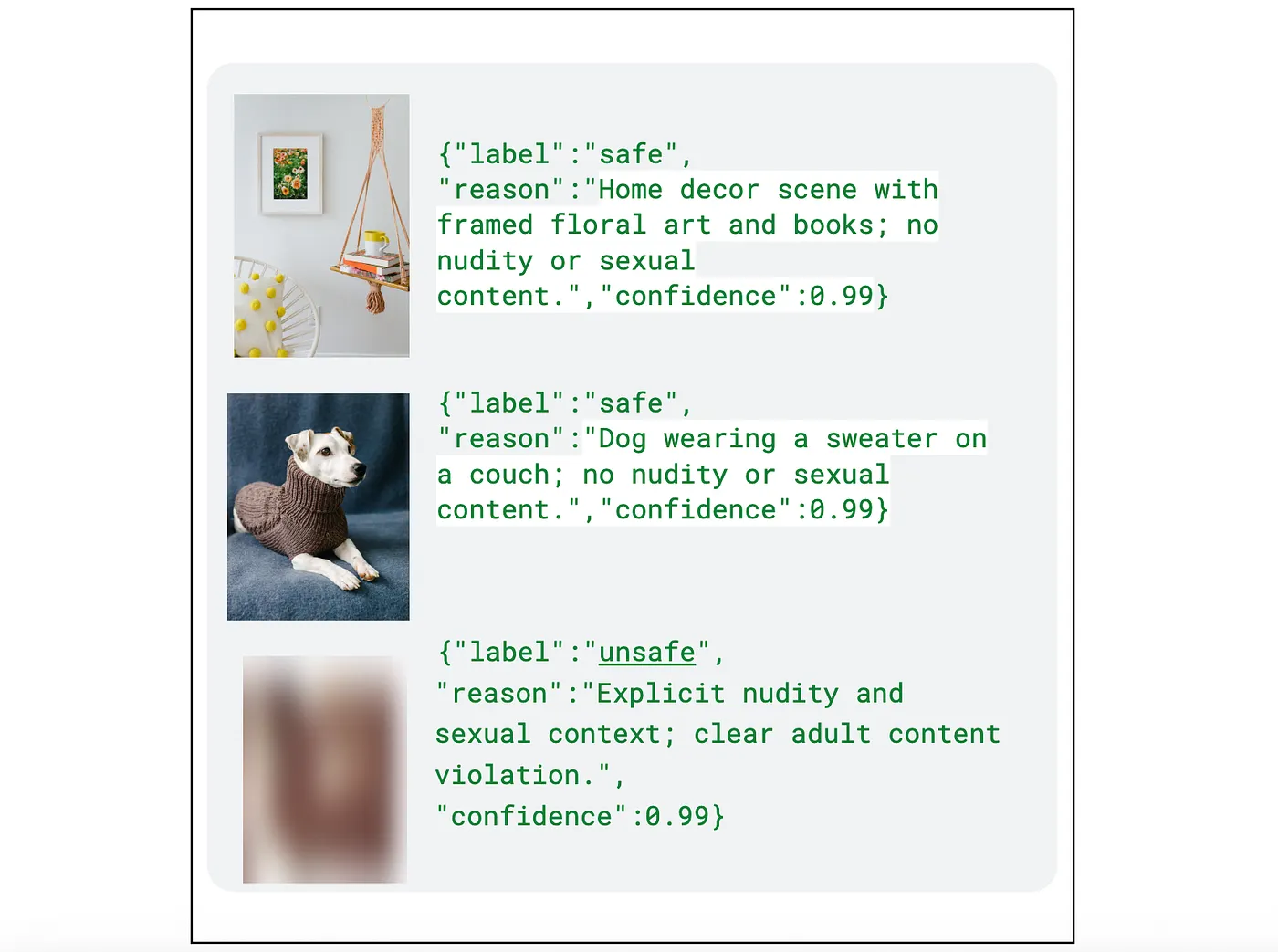}
 \caption{Example LLM outputs: structured label, rationale, and confidence. (Content blurred.)}
 \Description{Three example images with JSON-like LLM outputs marking safe or unsafe with brief rationales.}
 \label{fig:llm_examples}
\end{figure}

Given the sampled content IDs, we label each unit with a multimodal LLM using (i) the content (image/video keyframes, title, text metadata), (ii) an SME-reviewed policy description, and (iii) an optimized prompt that returns a structured label and brief rationale.
Our deployment used GPT-4-class multimodal models and has since moved to current-generation models; outputs are structured JSON-like records (label, reason, confidence).
Because rare-event prevalence bias is dominated by labeler FPR (\S\ref{sec:labelerror}), prompt optimization explicitly targets FPR minimization subject to recall constraints.
Figure~\ref{fig:llm_examples} shows example outputs.

\subsection{Quality gating and monitoring}
\label{sec:gating}
Before launching a policy metric, the LLM+prompt configuration must meet a minimum decision-quality bar on a gold set labeled by policy SMEs (ground truth).
Concretely, LLM prevalence measurements are validated against $3\times$ majority-vote human labeling, and the metric ships only when the LLM-vs-human bias and decision-quality deltas are within policy-specific tolerance bounds.
After launch, we continuously monitor decision quality using two mechanisms:
(1) a random subsample of daily items routed to human validation (drawn from the same probability design, see \S\ref{sec:labelerror}), and
(2) periodic re-evaluation on refreshed gold sets to detect drift (policy drift, content shift, or model updates).
This governance mirrors best practices for aligning automated evaluators with human judgment \cite{gilardi2023chatgpt,zheng2023judging,hashemi2024llmrubric} and is detailed in a companion decision-quality framework paper \cite{tian2026dqframework}.

In internal evaluations on SME-validated gold sets, the LLM-based labeler achieved accuracy within $\sim$3.2\% (relative difference) and an F1 score within $\sim$0.5\% (relative difference) of human agent performance \cite{tian2026dqframework}.
Appendix~\ref{app:dq} reports per-metric deltas (precision, recall, F1, FPR, FNR, informedness) for individual frontier models and a $3\times$ LLM majority ensemble against a single-review human baseline across two policy areas: the LLM panel improves recall and FPR simultaneously over typical single-model operation while recovering most of the gap to single-human review, and even a $3\times$ human majority differs measurably from a single review.
These results demonstrate that the automated pipeline performs on par with human review while providing the scalability required for daily measurement.

\section{System Implementation and Configurable Workflow}
\label{sec:system}
\begin{figure*}
 \centering
 \includegraphics[width=0.98\textwidth]{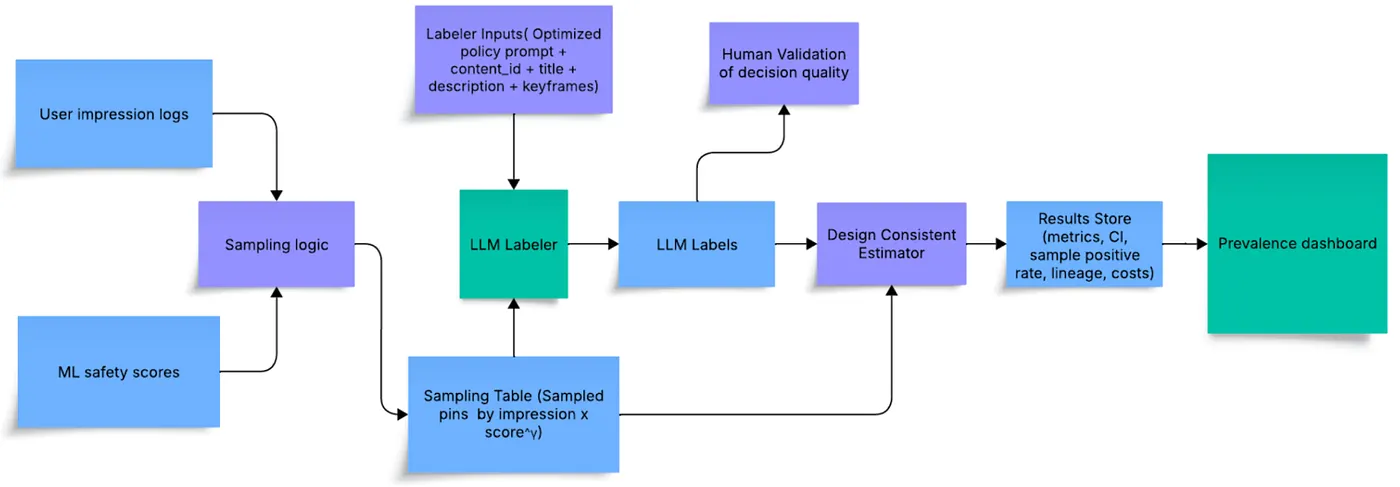}
 \caption{Prevalence workflow: impression logs and auxiliary safety scores drive sampling; an LLM labeler produces policy labels; the estimator generates prevalence and CIs; results and lineage are stored for dashboards and audits.}
 \Description{A block diagram showing impression logs and ML scores feeding sampling logic; sampled items go to an LLM labeler; labels feed an estimator; results are stored and shown in a dashboard, with a human validation loop.}
 \label{fig:system}
\end{figure*}

The engineering realization follows three principles: metric definitions are declarative configuration rather than bespoke pipelines; every run persists full lineage so any published number can be audited; and cost controls are first-class because labeling spend scales with the daily budget $m$.
Figure~\ref{fig:system} summarizes the end-to-end workflow.

\subsection{Config-driven measurement}
A key engineering principle is that prevalence measurement is \emph{configurable}.
Each metric definition specifies:
(1) the target policy taxonomy (including sub-policies),
(2) data sources for impressions and auxiliary scores,
(3) sampling parameters $(m,\nu,\gamma,\epsilon)$,
(4) the LLM prompt and model identifiers,
(5) required gold sets and quality thresholds,
and (6) output schemas and dashboards.
This abstraction allows teams to stand up new policy measurements quickly, following the onboarding workflow of \S\ref{sec:tuning}.

\subsection{Lineage, auditability, and cost controls}
\label{sec:lineage}
For each daily run we persist:
\textbf{(i)} sampled IDs with weights and segment breakdowns,
\textbf{(ii)} LLM outputs with prompt/policy/model versions,
\textbf{(iii)} estimator diagnostics (CI width, ESS, positive rate), and
\textbf{(iv)} token usage and per-run cost.
Compared with a human-only workflow that can take weeks to reach statistical significance, LLM-assisted labeling can produce high-quality labels within hours and at orders-of-magnitude lower cost \cite{pinterest_radar_2025}.
Daily label budgets $m$ balance precision against run-time and token cost: larger $m$ tightens CIs but increases spend and can challenge a daily refresh SLA.
Prompt and policy updates are gated by SME review and gold-set thresholds, adding governance overhead but preventing silent metric shifts.
As part of measurement governance, when updating the auxiliary-score model used for sampling we run a consistency check using both the new and previous score versions (and, optionally, $\gamma=0$) and confirm that point estimates agree within statistical uncertainty; differences should mainly appear in CI width through weight variability.
Because the LLM labeler is independent of enforcement models (which are used only as auxiliary scores), the resulting measurement is less coupled to the production thresholds.

\subsection{Post-launch operational efficiency}
The integration of LLM labeling provides a step-function increase in measurement velocity. Our automated system processes over 1M units per day, representing a throughput increase of approximately 100$\times$ that of a human baseline. This efficiency allows for daily measurements, which was previously impossible due to latency.

\begin{table}[h]
  \caption{Relative Operational Comparison: Human vs. LLM}
  \label{tab:efficiency}
  \begin{tabular}{lll}
    \toprule
    Metric & Human Review & LLM\\
    \midrule
    Label Throughput & Baseline ($1\times$) & $\sim 100\times$ \\
    Measurement Frequency & Monthly/Quarterly & Daily \\
    Labeling Latency & Days to Weeks & $<$ 24 Hours \\
    Cost per 1k labels & Baseline & $\sim 95\%$ reduction \\
    \bottomrule
  \end{tabular}
\end{table}

Our production configuration uses roughly 3,000 tokens for the policy prompt and 400 per image (keyframes and metadata), with at most 100 output tokens for the rationale; batch API processing and prompt caching reduce these costs further.

\section{Empirical Results}
\label{sec:eval}
This section provides compact quantitative evidence for four key claims: (i) ML-assisted probability sampling improves sampling efficiency for rare policy areas at a fixed label budget, (ii) design-based reweighting keeps point estimates design-consistent under auxiliary-score drift, which surfaces as variance rather than bias, (iii) the \S\ref{sec:tuning} hyperparameter trade-off is well behaved in a realistic range but degrades under aggressive tilting, and (iv) a single global probability sample supports multiple segment pivots with usable uncertainty.
We report results from a simulation POC plus anonymized production statistics in relative terms.

\paragraph{Setup.}
We simulate a large population with long-tailed impression weights, rare violations, and a noisy auxiliary risk score correlated with the true label.
We compare two schemes at a fixed daily label budget $m$:
\textbf{(a)} PPS with probabilities proportional to impressions only ($\gamma=0$), and
\textbf{(b)} ML-assisted PPS with probabilities proportional to impressions times a score transform ($\gamma=1$).
For each scheme we estimate impression-weighted prevalence using a Hansen--Hurwitz estimator and report empirical 95\% uncertainty over repeated trials.
As expected for a design-based estimator, empirical bias is negligible across sample sizes $m$ (not shown).
Implementation details and the exact synthetic parameter settings used to generate Figure~\ref{fig:ci_width_vs_m} are provided in Appendix~\ref{app:sim_pseudocode}.

\paragraph{Sampling efficiency.}
Figure~\ref{fig:ci_width_vs_m} shows CI width versus label budget.
At realistic sample sizes, ML-assisted PPS yields materially tighter CIs than uniform sampling and PPS-by-impressions, consistent with concentrating budget on high-exposure, high-risk items to increase effective positives for rare events.

\begin{figure}[htbp]
 \centering
 \includegraphics[width=0.8\linewidth]{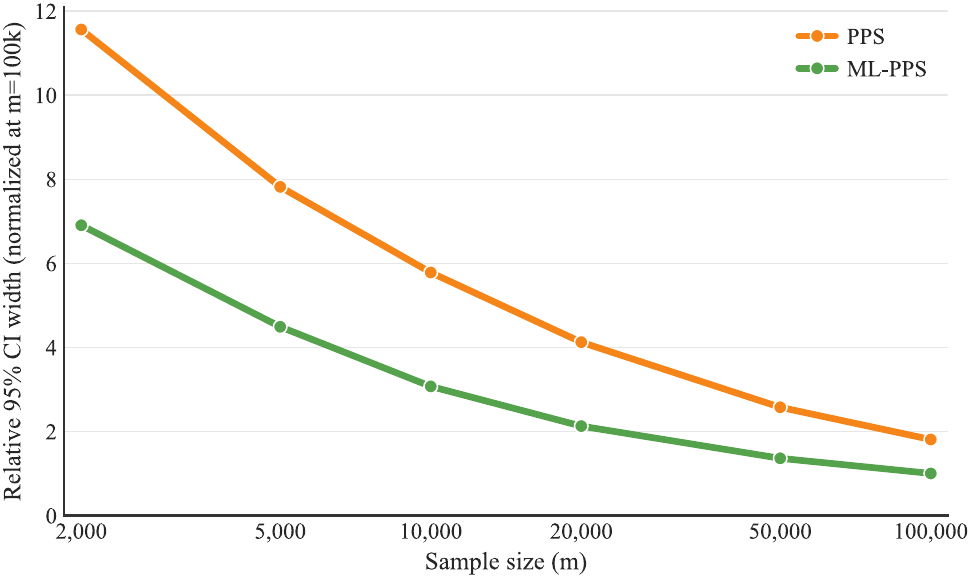}
 \caption{Simulation POC: empirical 95\% CI width vs. sample size $m$ for two sampling schemes: PPS ($w=\text{impressions}$) and ML-PPS ($w=\text{impression} \times \text{model\_score}$)}
 \label{fig:ci_width_vs_m}
\end{figure}

\paragraph{Hyperparameter sensitivity ($\gamma$).}
Using the same synthetic population at a fixed budget $m=10{,}000$, we sweep the score exponent $\gamma$ (Appendix~\ref{app:hyper}, Figure~\ref{fig:gamma}).
Moving from $\gamma=0$ to $\gamma=1$ narrows the empirical CI width by roughly $46\%$ as the sample positive fraction rises from $0.5\%$ to $1.9\%$; $\gamma=2$ buys a further $\sim$15\% but collapses $\mathrm{ESS}/m$ to $0.08$, and $\gamma=4$ backfires outright (CI $2.9\times$ wider than $\gamma=1$, $\mathrm{ESS}/m\approx 0.01$).
This supports the recommendation of \S\ref{sec:tuning}: use the smallest $\gamma$ that meets the positive-count target, keeping an ESS margin.

\paragraph{Score drift (synthetic).}
We then degrade the auxiliary score's class separation while keeping labels fixed, holding $\gamma=1$ (Appendix~\ref{app:hyper}, Table~\ref{tab:drift}).
Across no-, moderate-, and severe-drift conditions, the empirical bias stays below $0.3\%$ relative (within Monte Carlo noise), while the empirical CI widens by $+27\%$ and $+63\%$ respectively as the positive fraction falls from $1.9\%$ to $0.8\%$; Kish ESS stays roughly flat, which is why dashboards monitor positive counts alongside ESS.
In production, every auxiliary-model refresh runs the dual-scoring consistency check of \S\ref{sec:lineage}; observed differences have appeared as modest CI-width changes rather than level shifts, consistent with the simulation.

\paragraph{Production sampling efficiency (relative).}
We also evaluate sampling efficiency in production over one year of daily runs by comparing impression-only PPS (\S\ref{sec:estimation}) to ML-assisted sampling (ML-PPS; Eq.~\ref{eq:weight}) at a fixed label budget.
Across multiple policy areas, ML-assisted sampling increases the sample positive fraction by $6$--$11\times$ (Table~\ref{tab:lift}), reducing the label budget required to reach a target CI width.
Absolute values are omitted for confidentiality; estimates remain design-consistent because they reweight by known sampling probabilities.

\begin{table}
 \caption{Production: lift in sample positive fraction from ML-assisted sampling (ML-PPS) relative to impression-only PPS at fixed label budget. Values are ratios (higher is better).}
 \label{tab:lift}
 \begin{tabular}{lc}
 \toprule
 Policy area & Positive-rate lift (ML-PPS : PPS) \\
 \midrule
 A & 6.0$\times$ \\
 B & 11.4$\times$ \\
 C & 10.8$\times$ \\
 \bottomrule
 \end{tabular}
\end{table}

\section{Extensions: Experimentation and User Reach}
Beyond monitoring, prevalence supports decision-making for interventions.

\paragraph{Experimentation.}
To measure treatment effects on rare policy areas with acceptable variance, we combine probability sampling for unbiased estimation with calibrated score-to-risk mappings, enabling surrogate prevalence metrics for large-scale A/B testing \cite{xu2026surrogateprevalence}: estimate a calibration map from an unbiased labeled sample, then apply it to larger impression logs to form lower-variance experiment metrics.

\paragraph{User reach.}
A complementary metric is \emph{user reach}: the fraction of daily active users who saw at least one violative impression. It requires user-level aggregation and can be approached via a two-stage design (sample users, then impressions within users) or model-assisted approximations; we leave a full treatment to future work.

\section{Case Study: Sensitivity, Minimum Detectable Effects, and Alerting}
\label{sec:case}
Two questions recur in production: (i) \emph{How sensitive is the metric?} and (ii) \emph{How large a change can we reliably detect?}
Because prevalence comes from a probability sample, sensitivity is a tunable knob: increase daily label volume (or improve weight efficiency) to tighten CIs.
For time aggregation we now make the day index $d$ explicit.

\subsection{Sensitivity as CI half-width}
We define daily \emph{sensitivity} as the half-width of the 95\% CI,
\begin{equation}
 h_d \;=\; 1.96\,\sqrt{\widehat{\mathrm{Var}}(\hat{\theta}_d)}.
\end{equation}
When violations are rare and weights are not too variable, a useful back-of-the-envelope approximation relates sensitivity to an effective sample size (ESS):
\begin{equation}
 h_d \;\approx\; 1.96\,\sqrt{\frac{\theta_d(1-\theta_d)}{\mathrm{ESS}_d}}
 \;\approx\; 1.96\,\sqrt{\frac{\theta_d}{\mathrm{ESS}_d}}\quad(\theta_d\ll 1).
\label{eq:sens_ess}
\end{equation}
Sensitivity thus scales roughly as $1/\sqrt{\mathrm{ESS}_d}$: doubling label volume improves CI width by about $\sqrt{2}$, and for ML-assisted designs $\mathrm{ESS}_d$ can be materially smaller than $m$ when inverse-probability weights are heavy-tailed.

\subsection{Example daily configurations}
Table~\ref{tab:monitoring} summarizes representative production configurations across policy areas.
All use ML-assisted sampling (Eqs.~(\ref{eq:weight})--(\ref{eq:key})) and LLM labeling (\S\ref{sec:llm}), with daily label budgets tuned per policy following \S\ref{sec:tuning}.
In practice, $m$ is increased for rarer types, deeper drill-downs, or when CI widths exceed an operational target.

\begin{table}
 \caption{Typical minimum detectable effects (MDE), relative to each policy area's mean. MDEs are computed for a two-sided test with $\alpha = 0.05$ and 80\% power. Policy areas A, B, and C differ in production prevalence and ML model quality. We omit absolute prevalence and quality values. Representative configurations; actual production parameters may vary.}
 \label{tab:monitoring}
 \begin{tabular}{lcc}
 \toprule
 Policy area & Daily labels $m$ & MDE (relative to mean) \\
 \midrule
 A & 50k & 6.0\% \\
 B & 100k & 6.5\% \\
 C & 10k & 5.0\% \\
 \bottomrule
 \end{tabular}
\end{table}

The system runs as a daily batch pipeline: each run draws fixed-size probability samples and produces six-figure-scale LLM-assisted multimodal labels under ongoing decision-quality monitoring.
We next translate daily sensitivity into weekly minimum detectable effects (MDEs) used for alerting.

\subsection{From daily sensitivity to weekly MDE}
\label{sec:mde}
For operational alerting we smooth the daily series with a 7-day moving average,
\begin{equation}
 \bar\theta_d \;=\; \frac{1}{7}\sum_{t=d-6}^{d} \hat\theta_t.
\label{eq:ma7}
\end{equation}
If daily estimation noise is approximately independent with variance $\sigma^2$, then
$\mathrm{Var}(\bar\theta_d)\approx \sigma^2/7$ and the standard error shrinks by $\sqrt{7}$.
To detect a step change by comparing two (approximately independent) 7-day windows at two-sided level $\alpha$ and power $1-\beta$, a standard normal approximation yields
\begin{equation}
 \mathrm{MDE}_{\mathrm{abs}} \;\approx\; \bigl(z_{1-\alpha/2}+z_{1-\beta}\bigr)\,\sqrt{\frac{2}{7}}\,\sigma.
\label{eq:mde_abs}
\end{equation}
Expressed as a relative change from a baseline $\theta_0$ and using $h_d \approx z_{1-\alpha/2}\sigma$,
\begin{equation}
 \mathrm{MDE}_{\mathrm{rel}} \;\approx\; \frac{\mathrm{MDE}_{\mathrm{abs}}}{\theta_0}
 \;\approx\; \underbrace{\frac{z_{1-\alpha/2}+z_{1-\beta}}{z_{1-\alpha/2}}\sqrt{\frac{2}{7}}}_{\approx\,0.76\ \text{for }\alpha=0.05,\ 1-\beta=0.8}\,\frac{h_d}{\theta_0}.
\label{eq:mde_rel}
\end{equation}
Equation~\eqref{eq:mde_rel} highlights that relative detectability scales with the ratio $h_d/\theta_0$ and benefits from temporal aggregation (e.g., weekly rollups).

\paragraph{Correlated fluctuations.}
Real systems exhibit autocorrelation due to content mix shifts, policy changes, and labeler drift.
If $\hat\theta_d$ has lag-$\ell$ correlation $\rho_\ell$, then
$\mathrm{Var}(\bar\theta_d)$ is inflated by
$1+2\sum_{\ell=1}^{6}(1-\ell/7)\rho_\ell$.
We therefore treat \eqref{eq:mde_rel} as a best-case planning rule and validate MDEs empirically from historical residuals.

\subsection{What dominates short-term fluctuations}
In many policy areas, day-to-day variability is driven less by sampling error and more by (i) \emph{decision quality} (prompt/policy interpretation, ambiguous edge cases) and (ii) \emph{topic mix shifts} (changes in the distribution of content types and contexts).
We mitigate these effects through (a) decision-quality gates and recurring gold-set evaluations (\S\ref{sec:llm}), (b) drift diagnostics (auxiliary score distributions, topic/surface composition), and (c) smoothing for alerting (\eqref{eq:ma7}).
When decision quality is the limiting factor, increasing $m$ alone does not improve metric conviction; instead, prompt iteration and targeted human validation are required.

In addition, because the estimator is exposure-weighted, a small number of high-weight sampled units can disproportionately affect the daily estimate, particularly when the underlying rate is very low; the bootstrap influence diagnostics of \S\ref{sec:labelerror} flag such days automatically.
Highly impressed positives that recur across consecutive daily samples receive dedicated study: SME adjudication classifies each as (i) an area of policy ambiguity, feeding policy clarification and prompt updates, (ii) an emerging threat vector, feeding enforcement and monitoring, or (iii) a genuine false detection, feeding gold sets and prompt optimization as an RLHF-style SME feedback signal.
Because every run persists the intermediary tables (sampled units, weights, segment breakdowns, labels, and rationales; \S\ref{sec:lineage}), this root-cause analysis is a query over lineage rather than a bespoke study.

\section{Operational Impact}
\label{sec:impact}
AI-assisted prevalence measurement transforms safety monitoring from infrequent audits into a high-frequency, statistically grounded feedback loop: proactive detection of emerging risks with fast root-cause analysis via drill-downs, quicker evaluation of launches and mitigations, and scalable governance through versioned prompts, gold sets, and lineage that reserve human review for targeted validation.

\section{Discussion and Limitations}
\textbf{Rare policy areas.} Very low base rates lead to wide daily CIs. We adapt by tuning $(\nu,\gamma)$, increasing $m$, stratifying, or pooling to weekly windows.

\textbf{Prompt and policy drift.} Policy definitions evolve and content patterns shift. We version policies/prompts and periodically backfill labels for comparability.

\textbf{Labeler drift.} LLM providers update models; we mitigate with pinned model versions when possible and ongoing validation. The double-digit recall spread across individual models in Appendix~\ref{app:dq} shows why gold-set gating on every model update is non-negotiable.

\textbf{Decision quality as a first-class object.} Our label-error analysis (\S\ref{sec:labelerror}) treats $(r,f)$ as scalar summaries; richer treatment of labeler behavior (per-sub-policy error structure, prompt optimization, cost-quality frontiers) is the subject of a companion paper \cite{tian2026dqframework} and planned follow-up work.

\textbf{Confidentiality constraints.} Absolute prevalence levels and policy-specific model quality remain confidential; we compensate with fully specified synthetic studies (Appendices~\ref{app:sim_pseudocode}, \ref{app:label_error}, \ref{app:hyper}) so the methodology is independently verifiable.

\section{Ethical Considerations}
\label{sec:ethics}
\textbf{Reviewer wellbeing.} The LLM performs bulk labeling; human review is reserved for targeted validation by trained professionals under established wellness safeguards, minimizing human exposure to harmful content.

\textbf{Measurement, not enforcement.} Labels from this pipeline are used exclusively for aggregate measurement: they trigger no enforcement actions against content or users, and the labeler is decoupled from enforcement models (\S\ref{sec:lineage}). Measurement complements, and does not substitute for, policy, legal, and trust operations, which remain governed by established internal processes and external regulatory obligations.

\textbf{Privacy and data governance.} Sampling operates on content-level impression aggregates; labeling inputs are content and metadata, not user identities. Samples, labels, and lineage live in access-controlled tables with retention limits; absolute levels for sensitive policies are withheld per regulatory and legal obligations.

\textbf{Fairness and error monitoring.} Segment drill-downs from one unbiased sample surface unevenly distributed harms, and monitoring $(r,f)$ on design-consistent validation samples guards against systematic labeler bias across content types.

\section{Conclusion}
We presented a prevalence measurement system for platform safety combining ML-assisted probability sampling with LLM labeling: design-consistent daily estimates with confidence intervals, many drill-downs from a single sample, and a configurable, lineage-rich implementation that expands to new policies under gold-set governance, with failure modes made explicit by the label-error and hyperparameter sensitivity analyses.

\bibliographystyle{ACM-Reference-Format}
\bibliography{kdd_prevalence_llm_extended}

\appendix
\section{Simulation details for Figure~\ref{fig:ci_width_vs_m}}
\label{app:sim_pseudocode}
This appendix provides a reproducible description of the simulation POC used in \S\ref{sec:eval} to generate Figure~\ref{fig:ci_width_vs_m}.
The setup is deliberately simple: its goal is not to mirror any specific production policy area, but to stress-test relative sampling efficiency under three conditions we consistently observe in practice: (i) heavy-tailed impression counts, (ii) rare violations, and (iii) an imperfect but useful auxiliary score correlated with the true label.
All parameters are synthetic and are not fit to any production dataset.
The same population (identical parameters and seeds) is reused for the label-error study of Appendix~\ref{app:label_error} and the hyperparameter and drift studies of Appendix~\ref{app:hyper}.

\begin{algorithm}[H]
\caption{Simulation POC used for Figure~\ref{fig:ci_width_vs_m}}
\label{alg:ciwidth_vs_m}
\begin{algorithmic}[1]
\Require population size $N$, base positive rate $p$, sample sizes $\mathcal{M}$, trials $T$
\Require impression parameters $(p_{\text{small}}, \alpha, x_m)$; score parameters $(a_-, b_-), (a_+, b_+)$
\Require sampling hyperparameters $(\nu, \gamma, \epsilon)$ and random seeds $(seed_{\text{pop}}, seed_{\text{mc}})$
\State Set RNG seed to $seed_{\text{pop}}$ and generate a synthetic population of $N$ items:
\For{$i=1$ to $N$}
 \State Sample label $y_i \sim \mathrm{Bernoulli}(p)$
 \State Sample impressions $w_i$ from mixture:
 \State \hspace{1em}with prob $p_{\text{small}}$: $w_i \sim \mathrm{Unif}\{1,\dots,10\}$
 \State \hspace{1em}else: $w_i \leftarrow x_m\bigl(1+\mathrm{Pareto}(\alpha)\bigr)$
 \State Sample auxiliary score $s_i$ from $\mathrm{Beta}(a_-, b_-)$ if $y_i=0$, else $\mathrm{Beta}(a_+, b_+)$
\EndFor
\State Define sampling distributions (with replacement):
\State \hspace{1em}PPS: $\pi_i^{(\mathrm{pps})} \propto w_i^{\nu}$
\State \hspace{1em}ML-PPS: $\pi_i^{(\mathrm{ml})} \propto w_i^{\nu}\bigl(s_i^{\gamma} + \epsilon\bigr)$
\State Set RNG seed to $seed_{\text{mc}}$ for Monte Carlo sampling.
\For{each $m \in \mathcal{M}$}
 \For{each scheme $b \in \{\mathrm{pps}, \mathrm{ml}\}$}
 \For{$t=1$ to $T$}
 \State Draw $m$ items i.i.d.\ with replacement using $\pi^{(b)}$
 \State Compute Hansen--Hurwitz ratio estimate:
 \State \hspace{1em}$\hat{\theta}^{(t)}_{m,b} =
 \dfrac{\frac{1}{m}\sum_{j} y_{j}w_{j}/\pi^{(b)}_{j}}{\frac{1}{m}\sum_{j} w_{j}/\pi^{(b)}_{j}}$
 \EndFor
 \State Empirical CI width: $W_{m,b} \leftarrow Q_{0.975}(\hat{\theta}_{m,b}) - Q_{0.025}(\hat{\theta}_{m,b})$
 \EndFor
\EndFor
\State Normalize for plotting: $W^{\mathrm{rel}}_{m,b} \leftarrow W_{m,b}/W_{100{,}000,\mathrm{ml}}$
\end{algorithmic}
\end{algorithm}

\begin{table}[t]
 \centering
 \caption{Synthetic simulation parameters used to generate Figure~\ref{fig:ci_width_vs_m}. All parameters are hypothetical and not fit to any production dataset.}
 \label{tab:sim_params_fig4}
 \footnotesize
 \begin{tabular}{@{}lp{0.62\linewidth}@{}}
 \toprule
 Parameter & Value \\
 \midrule
 Population size $N$ & $300{,}000$ items \\
 Base positive rate $p$ & $0.005$ (0.5\%) \\
 Impressions mixture mass $p_{\text{small}}$ & $0.93$ \\
 Impressions small component & $\mathrm{Unif}\{1,\dots,10\}$ \\
 Impressions tail component & $x_m(1+\mathrm{Pareto}(\alpha))$, $\alpha=1.4$, $x_m=10$ \\
 Scores $s \mid y=0$ & $\mathrm{Beta}(1.5, 6.0)$ \\
 Scores $s \mid y=1$ & $\mathrm{Beta}(6.0, 1.5)$ \\
 Sampling weights (Eq.~\ref{eq:weight}) & $\nu=1.0$; $\gamma=0$ (PPS) or $\gamma=1$ (ML-PPS); $\epsilon=10^{-6}$ \\
 Sample sizes $\mathcal{M}$ & $\{2{,}000, 5{,}000, 10{,}000, 20{,}000, 50{,}000, 100{,}000\}$ \\
 Monte Carlo trials $T$ & $500$ per $(m,\mathrm{scheme})$ \\
 Random seeds & population generation: $seed_{\text{pop}}=42$; sampling draws: $seed_{\text{mc}}=123$ \\
 \bottomrule
 \end{tabular}
\end{table}

\section{Decision quality versus a single-review human baseline}
\label{app:dq}
Table~\ref{tab:dq} substantiates the human-vs-LLM decision-quality claims of \S\ref{sec:gating}.
All values are percentage-point deltas relative to a single review by a trained, non-expert professional reviewer (1$\times$ BPO), evaluated against SME-adjudicated gold labels, for two policy areas.
Informedness is Youden's $J$ (recall $-$ FPR); for FPR and FNR, lower (more negative $\Delta$) is better.
Model names refer to point-in-time versions evaluated during the deployment window.

Three observations stand out.
First, even a $3\times$ human majority vote improves materially on a single review (F1 $+3.4$ and $+2.7$ points), quantifying the noise floor of the single-review baseline itself and motivating the $3\times$ majority-vote validation gate of \S\ref{sec:gating}.
Second, the best individual frontier models trail the single-review baseline modestly, and a $3\times$ LLM majority vote recovers most of the remaining gap (F1 within $1.8$ and $0.6$ points of the baseline), with FPR at or below the human baseline in the tuned policy area A configuration, consistent with the FPR-minimization objective motivated in \S\ref{sec:labelerror}.
Third, a panel of LLM judges can move decision quality in a Pareto-positive direction out of the box: by outvoting idiosyncratic errors in both directions, the $3\times$ majority improves recall and lowers FPR simultaneously relative to single-model operation, rather than merely trading precision against recall; in policy area A it matches the lowest constituent FPR while beating the recall of both low-FPR members.
Beyond gating, the accumulated SME adjudications could further be leveraged for RLHF-style improvement of labeler decision quality.

\begin{table*}[t]
 \centering
 \caption{Decision-quality deltas (percentage points) versus a $1\times$ single non-expert human review (BPO) baseline per policy area, scored against SME-adjudicated gold sets. Column arrows mark the better direction: higher recall is better, lower FPR is better. ``(high)'' denotes the high-reasoning setting; the ``full context'' configuration supplies the complete policy document as labeling context. Policy area A: gold set after one round of SME re-adjudication; prompts optimized to minimize FPR. Policy area B: production prompt prior to SME review. $3\times$ majority rows aggregate three independent human reviews or three independent LLM calls. Informedness $=$ recall $-$ FPR.}
 \label{tab:dq}
 \footnotesize
 \begin{tabular}{@{}lrrrrrr@{\hspace{2.2em}}lrrrrrr@{}}
 \toprule
 \multicolumn{7}{c}{Policy area A} & \multicolumn{7}{c}{Policy area B} \\
 \cmidrule(r{2em}){1-7}\cmidrule(l){8-14}
 Labeler & $\Delta$Prec.$\uparrow$ & $\Delta$Rec.$\uparrow$ & $\Delta$F1$\uparrow$ & $\Delta$FPR$\downarrow$ & $\Delta$FNR$\downarrow$ & $\Delta$Inf.$\uparrow$ &
 Labeler & $\Delta$Prec.$\uparrow$ & $\Delta$Rec.$\uparrow$ & $\Delta$F1$\uparrow$ & $\Delta$FPR$\downarrow$ & $\Delta$FNR$\downarrow$ & $\Delta$Inf.$\uparrow$ \\
 \midrule
 1$\times$ BPO (baseline) & 0.0 & 0.0 & 0.0 & 0.0 & 0.0 & 0.0 &
 1$\times$ BPO (baseline) & 0.0 & 0.0 & 0.0 & 0.0 & 0.0 & 0.0 \\
 3$\times$ BPO majority & $+3.4$ & $+3.3$ & $+3.4$ & $-1.9$ & $-3.3$ & $+5.1$ &
 3$\times$ BPO majority & $+2.6$ & $+2.8$ & $+2.7$ & $-1.3$ & $-2.8$ & $+4.2$ \\
 GPT-5 (baseline prompt) & $-8.6$ & $-8.4$ & $-8.5$ & $+4.5$ & $+8.4$ & $-12.9$ &
 & & & & & & \\
 GPT-5.4 (high) & $+1.7$ & $-8.0$ & $-3.8$ & $-2.7$ & $+8.0$ & $-5.3$ &
 GPT-5.4 (high) & $-0.2$ & $-7.5$ & $-4.5$ & $-0.6$ & $+7.5$ & $-6.9$ \\
 GPT-5.4 (full context) & $-2.3$ & $+2.0$ & $+0.2$ & $+2.3$ & $-2.0$ & $-0.3$ &
 & & & & & & \\
 Gemini-3.1-pro & $-4.3$ & $+4.0$ & $+0.2$ & $+3.3$ & $-4.0$ & $+0.7$ &
 Gemini-3.1-pro & $-3.6$ & $+8.2$ & $+2.4$ & $+3.7$ & $-8.2$ & $+4.5$ \\
 Claude Opus-4.6 & $+1.7$ & $-14.0$ & $-8.8$ & $-2.7$ & $+14.0$ & $-11.3$ &
 Claude Opus-4.6 & $-3.9$ & $-4.1$ & $-4.1$ & $+2.1$ & $+4.1$ & $-6.3$ \\
 3$\times$ LLM majority & $+3.7$ & $-5.0$ & $-1.8$ & $-2.7$ & $+5.0$ & $-2.3$ &
 3$\times$ LLM majority & $-0.8$ & $-0.4$ & $-0.6$ & $+0.6$ & $+0.4$ & $-1.0$ \\
 \bottomrule
 \end{tabular}
\end{table*}

\section{Label-error sensitivity study}
\label{app:label_error}
This appendix expands the label-error analysis of \S\ref{sec:labelerror}.

\paragraph{Heterogeneous, weight-correlated errors.}
Let each item $j$ have its own error rates: sensitivity $r_j=\Pr(\hat Y_j=1\mid Y_j=1)$ and false positive rate $f_j=\Pr(\hat Y_j=1\mid Y_j=0)$, where $\hat Y_j$ is the LLM label.
Replacing $Y_{J_i}$ with $\hat Y_{J_i}$ in the HH ratio estimator \eqref{eq:hh_ratio}, the estimand becomes
\begin{equation}
\theta^{(L)}
= \frac{\sum_j C_j\,\bigl[r_j Y_j + f_j (1-Y_j)\bigr]}{\sum_j C_j}
= \bar r\,\theta + \bar f\,(1-\theta),
\label{eq:weighted_rates}
\end{equation}
where
\begin{equation}
\bar r = \frac{\sum_{j: Y_j=1} C_j r_j}{\sum_{j: Y_j=1} C_j},
\qquad
\bar f = \frac{\sum_{j: Y_j=0} C_j f_j}{\sum_{j: Y_j=0} C_j}
\end{equation}
are the \emph{impression-weighted} error rates.
If labeler errors correlate with exposure or with the auxiliary risk score (e.g., high-risk, high-reach content is systematically harder), then $(\bar r,\bar f)$ differ from the unweighted error rates one would estimate on a convenience validation set, and both the monitored bias and any Rogan--Gladen correction would be misstated.
Estimating $(r,f)$ on validation subsamples drawn from the same probability design, and reweighting by the same $1/p_{J_i}$, yields design-consistent estimates of exactly $(\bar r,\bar f)$; this is the validation mechanism of \S\ref{sec:gating}.

\begin{figure}[t]
 \centering
 \includegraphics[width=\linewidth]{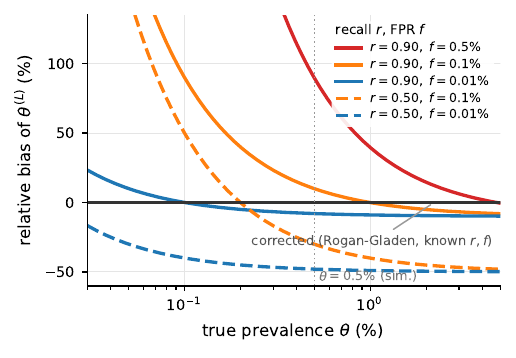}
 \caption{Relative bias of LLM-measured prevalence $\theta^{(L)}$ versus the true base rate $\theta$; color encodes the labeler false positive rate $f$, line style its recall $r$ (solid $r=0.90$, dashed $r=0.50$). At low prevalence the metric is extremely sensitive to FPR, which enters unbounded as $f/\theta$: at $\theta=0.1\%$, even $f=0.1\%$ makes about half of measured positives false detections ($+90\%$ relative at $r=0.90$). Recall loss enters only as a bounded offset: each dashed curve runs parallel to its solid counterpart, shifted down by a constant $40$ points. The solid horizontal line at zero marks the Rogan--Gladen corrected estimator with known $(r,f)$.}
 \Description{Line chart of relative bias in percent versus true prevalence on a log axis for five error profiles; false-positive-rate driven bias grows sharply as prevalence decreases, dashed low-recall curves run parallel to their solid counterparts at a constant offset, and a solid zero line marks the corrected estimator.}
 \label{fig:label_bias}
\end{figure}

\paragraph{Bias across base rates.}
Figure~\ref{fig:label_bias} plots the relative bias $(\theta^{(L)}-\theta)/\theta$ implied by \eqref{eq:measured} across base rates for representative error profiles.
The false positive rate dominates for rare policies: halving $r$'s distance to 1 changes the relative bias by a constant, while $f$ enters as $f/\theta$.
The curves cross zero at $\theta^\ast=f/(1-r+f)$, below which measured prevalence overstates and above which it understates the truth; with known $(r,f)$ the Rogan--Gladen correction removes the bias exactly.

\paragraph{Uncertainty propagation for the corrected estimator.}
Applying the delta method to \eqref{eq:rg} with independently estimated $(\hat r,\hat f)$ gives
\begin{equation}
\widehat{\mathrm{Var}}(\hat\theta_{\mathrm{RG}})
\approx
\frac{\widehat{\mathrm{Var}}(\hat\theta^{(L)})
 + \hat\theta_{\mathrm{RG}}^2\,\widehat{\mathrm{Var}}(\hat r)
 + (1-\hat\theta_{\mathrm{RG}})^2\,\widehat{\mathrm{Var}}(\hat f)}
{(\hat r-\hat f)^2}.
\label{eq:rg_var}
\end{equation}
For rare policies the $\widehat{\mathrm{Var}}(\hat f)$ term dominates: pinning down a small $f$ requires many validated \emph{negatives}, which is why our validation budgets are sized on the negative side and why we prefer monitoring $(\hat r,\hat f)$ over always-on correction when $\hat f\ll\hat\theta^{(L)}$ (\S\ref{sec:labelerror}).

\section{Hyperparameter and drift sensitivity}
\label{app:hyper}
Both studies reuse the synthetic population of Appendix~\ref{app:sim_pseudocode} (identical parameters, $seed_{\text{pop}}=42$) with $\nu=1$, $\epsilon=10^{-6}$, budget $m=10{,}000$, $T=500$ Monte Carlo trials, and sampling seed $seed_{\text{mc}}=123$; the true exposure-weighted prevalence of the realized population is $\theta=0.482\%$.

\begin{figure}[t]
 \centering
 \includegraphics[width=\linewidth]{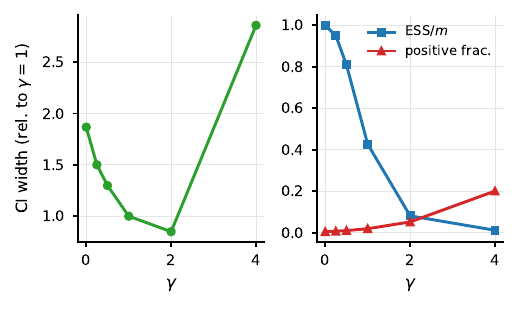}
 \caption{Sensitivity to the score exponent $\gamma$ at fixed budget $m=10{,}000$ on the Appendix~\ref{app:sim_pseudocode} population. Left: empirical 95\% CI width relative to $\gamma=1$. Right: Kish $\mathrm{ESS}/m$ and sample positive fraction. Aggressive tilting ($\gamma=4$) collapses ESS and widens the CI despite many positives.}
 \Description{Two small panels. Left panel shows CI width versus gamma forming a U shape with minimum near gamma between one and two. Right panel shows effective sample size over m decreasing with gamma while the positive fraction increases.}
 \label{fig:gamma}
\end{figure}

\paragraph{Sensitivity to $\gamma$.}
Figure~\ref{fig:gamma} sweeps $\gamma\in\{0, 0.25, 0.5, 1, 2, 4\}$.
CI width improves steeply from $\gamma=0$ to $\gamma=1$ as the sample positive fraction rises, is near-optimal on a broad plateau around $\gamma\in[1,2]$, and then degrades sharply: at $\gamma=4$ the empirical CI is $2.9\times$ wider than at $\gamma=1$ despite a $20\%$ positive fraction, because $\mathrm{ESS}/m$ collapses to about $0.01$ and a handful of extreme inverse-probability weights dominate every estimate.
Empirical bias is below $1\%$ relative for $\gamma\le 2$ and becomes unstable (mean deviation $+7.5\%$ over 500 trials) at $\gamma=4$, reflecting finite-sample ratio-estimator instability at tiny ESS rather than design bias.
This motivates the recommendation of \S\ref{sec:tuning}: choose the smallest $\gamma$ that meets the positive-count target, retaining an ESS margin.

\paragraph{Auxiliary-score drift.}
To emulate a stale score model, we redraw scores with degraded class separation while keeping labels and impressions fixed (drift seed $7$): moderate drift uses $s\mid y{=}1\sim\mathrm{Beta}(4.0,3.0)$, $s\mid y{=}0\sim\mathrm{Beta}(1.5,5.0)$; severe drift uses $\mathrm{Beta}(2.5,2.5)$ and $\mathrm{Beta}(1.6,4.0)$.
Table~\ref{tab:drift} confirms the design-based prediction: point estimates remain unbiased within Monte Carlo noise ($|$bias$|\le 0.3\%$ relative) while precision degrades, with CI width increasing $+27\%$ (moderate) and $+63\%$ (severe) as the positive fraction drops.
Notably, the Kish ESS on the denominator weights is nearly unchanged; the precision loss is driven by fewer positive draws, which is why production dashboards monitor positive counts alongside ESS (\S\ref{sec:tuning}).

\begin{table}[t]
 \centering
 \caption{Score drift experiment ($\gamma=1$, $m=10{,}000$, $T=500$): empirical relative bias, CI width relative to the no-drift condition, Kish $\mathrm{ESS}/m$, and mean sample positive fraction.}
 \label{tab:drift}
 \small
 \begin{tabular}{@{}lcccc@{}}
 \toprule
 Drift & Rel.\ bias & CI width (rel.) & $\mathrm{ESS}/m$ & Pos.\ frac. \\
 \midrule
 None & $-0.2\%$ & $1.00$ & $0.43$ & $1.9\%$ \\
 Moderate & $+0.0\%$ & $1.27$ & $0.43$ & $1.2\%$ \\
 Severe & $+0.2\%$ & $1.63$ & $0.49$ & $0.8\%$ \\
 \bottomrule
 \end{tabular}
\end{table}

% Optional: a compact visual schematic of the simulation pipeline (file: sim_poc_overview.pdf).
% Uncomment to include if space permits.
% \begin{figure}[t]
% \centering
% \includegraphics[width=\linewidth]{sim_poc_overview.pdf}
% \caption{Simulation pipeline used to generate Figure~\ref{fig:ci_width_vs_m}.}
% \label{fig:sim_poc_overview}
% \end{figure}

\end{document}